\pdfoutput=1

\documentclass[11pt]{article}

\usepackage{acl}

\usepackage{times}
\usepackage{latexsym}

\usepackage[T1]{fontenc}

\usepackage[utf8]{inputenc}

\usepackage{microtype}

\usepackage{hyperref}
\usepackage{xcolor}
\usepackage{tabularx}
\usepackage{multirow}
\usepackage{graphicx}
\usepackage{booktabs}
\usepackage{hyperref}
\usepackage{amsmath}
\usepackage{amsfonts}
\usepackage{multirow}
\usepackage{pgfplots}
\usepackage{pgfplotstable}
\usepackage{subcaption}
\newcommand{\abbr}{\textsc{HashEE}}

\title{A Simple Hash-Based Early Exiting Approach\\
For Language Understanding and Generation}

\author{
Tianxiang Sun\textsuperscript{\rm 1},
Xiangyang Liu\textsuperscript{\rm 1},
Wei Zhu\textsuperscript{\rm 2,3},
Zhichao Geng\textsuperscript{\rm 1},
Lingling Wu\textsuperscript{\rm 1},\\
\bf{
Yilong He\textsuperscript{\rm 3},
Yuan Ni\textsuperscript{\rm 3},
Guotong Xie\textsuperscript{\rm 3,4,5},
Xuanjing Huang\textsuperscript{\rm 1},
Xipeng Qiu\textsuperscript{\rm 1,6 \thanks{$^*$ Corresponding author (\texttt{xpqiu@fudan.edu.cn})}}
}\\
\textsuperscript{\rm 1}School of Computer Science, Fudan University
\quad \textsuperscript{\rm 2}East China Normal University\\
\textsuperscript{\rm 3}Ping An Health and Technology Company Limited \quad \textsuperscript{\rm 4}Ping An Healthcare Technology \\
\textsuperscript{\rm 5}Ping An International Smart City Technology Company Limited \quad \textsuperscript{\rm 6}Peng Cheng Laboratory\\
}

\begin{document}
\maketitle
\begin{abstract}


Early exiting allows instances to exit at different layers according to the estimation of difficulty.
Previous works usually adopt heuristic metrics such as the entropy of internal outputs to measure instance difficulty, which suffers from generalization and threshold-tuning. In contrast, learning to exit, or learning to predict instance difficulty is a more appealing way. Though some effort has been devoted to employing such "learn-to-exit" modules, it is still unknown whether and how well the instance difficulty can be learned. As a response, we first conduct experiments on the learnability of instance difficulty, which demonstrates that modern neural models perform poorly on predicting instance difficulty. Based on this observation, we propose a simple-yet-effective Hash-based Early Exiting approach (\abbr) that replaces the learn-to-exit modules with hash functions to assign each token to a fixed exiting layer. Different from previous methods, \abbr~requires no internal classifiers nor extra parameters, and therefore is more efficient. Experimental results on classification, regression, and generation tasks demonstrate that \abbr~can achieve higher performance with fewer FLOPs and inference time compared with previous state-of-the-art early exiting methods.

\end{abstract}

\section{Introduction}
Early exiting is a widely used technique to accelerate inference of deep neural networks. With the rising of pre-trained language models (PLMs)~\cite{Devlin2019BERT,Yang2019XLNet,Lan2020ALBERT,Raffel2020T5,Sun2020Colake,Qiu2020survey,Sun2021Paradigm}, early exiting is drawing increasing attention in the NLP community. At its core, early exiting allows simple instances to exit early while allowing hard instances to exit late. Thus, how to measure instance difficulty is a crucial problem.


Most existing early exiting methods attach multiple internal classifiers to the PLM and adopt some heuristic metrics, such as entropy~\cite{Xin2020DeeBERT,Liu2020FastBERT} or maximum softmax score~\cite{Schwartz2020Right} of internal outputs, to measure instance difficulty. However, these methods can not easily generalize to new tasks. On the one hand, these metrics are not accessible on some tasks such as regression. On the other hand, In order for these methods to perform well, one usually needs to fine-tune the threshold, which varies widely across different tasks and datasets.

Another way to measure instance difficulty is to directly learn it. Recent studies~\cite{Elbayad2020Depth,Xin2021BERxiT} that use the idea of "learn-to-exit" have achieved promising results. They jointly train a neural model to predict for each instance the exiting layer. At their core, the learn-to-exit module is to estimate the difficulty for each instance. Compared with previous heuristically designed metrics for difficulty, learn-to-exit is task-agnostic and does not require threshold-tuning, therefore is a more promising way.

Despite their success, it is still unknown whether or how well the instance difficulty can be learned. As a response, in this work, we construct datasets for two kinds of instance difficulty: (a) Human-defined difficulty, and (b) Model-defined difficulty. The dataset for human-defined difficulty has two labels, 0 for instances that can be annotated by human and 1 for instances that cannot. For model-defined difficulty, we train a multi-exit BERT~\cite{Devlin2019BERT}, which is attached with an internal classifier at each layer, on a sentence-level classification task, SNLI~\cite{Bowman2015SNLI}, and a token-level classification task, OntoNotes NER~\cite{Hovy2006Ontonotes}. The trained multi-exit BERTs are then used to annotate for each development instance whether it can be correctly predicted by each internal classifier. Thus, our constructed sentence-level and token-level model-defined difficulty datasets are multi-label classification datasets. Experimental results demonstrate that, modern neural networks perform poorly on predicting instance difficulty. This observation is consistent with previous work~\cite{Laverghetta2020Difficulty} on estimating instance difficulty for curriculum learning.

Given that instance difficulty is hard to be predicted, then what works in the learn-to-exit modules? We hypothesis that the consistency between training and inference may play an important role. That is, for a training instance $x_i$ that is predicted to exit at layer $l$, an inference instance $x_j$ that is similar with $x_i$ should be predicted to exit at layer $l$, too. Since neural networks are usually smooth functions~\cite{Ziegel2003Elements}, this consistency can be easily satisfied by neural learn-to-exit modules. If this hypothesis holds, we can replace the learn-to-exit module with a simple hash function. In particular, we use hash functions to assign each token to a fixed exiting layer. This hash-based early exiting method is named \abbr.

Compared with previous methods that use heuristic metrics for difficulty or jointly learn to exit, \abbr~offers several advantages: \textbf{(a)} \abbr~ requires no internal classifiers nor extra parameters, which are necessary in previous work. \textbf{(b)} \abbr~ can perform token-level early exiting without supervision, therefore can be widely used on various tasks including language understanding and generation. \textbf{(c)} The speed-up ratio can be easily tuned by modifying the hash function. \textbf{(d)} \abbr~can significantly accelerate model inference on a per-batch basis instead of per-instance basis as in previous work~\cite{Xin2020DeeBERT,Liu2020FastBERT,Zhou2020PABEE}.

We conduct experiments on classification, regression, and generation tasks. Experimental results on ELUE~\cite{Liu2021ELUE} demonstrate that \abbr, despite its simplicity, can achieve higher performance with fewer FLOPs and inference time than previous state-of-the-art methods on various tasks. Besides, our experiments on several text summarization datasets show that \abbr~can reduce $\sim$50\% FLOPs of BART~\cite{Lewis2020BART} and CPT~\cite{Shao2021CPT} while maintaining 97\% ROUGE-1 score.\footnote{Code is publicly available at \url{https://github.com/txsun1997/HashEE}.}


\section{Can Instance Difficulty Be Learned?}
In this section, we examine whether or to what extent instance difficulty can be learned. In particular, we manage to evaluate how well a neural network that trained on some data with difficulty annotation can generalize to unseen data. Here we consider two kinds of difficulty: human-defined difficulty and model-defined difficulty.

\subsection{Human-defined Difficulty}
\paragraph{Dataset Construction}
Human-defined difficulty of an instance measures how difficult for human to judge its label. To construct such a dataset, we use the SNLI dataset~\cite{Bowman2015SNLI}, which is a collection of 570k human-written English sentence pairs that are manually labeled with the inference relation between the two sentences: entailment, contradiction, or neutral. The labels in SNLI are determined by the majority of the crowd-sourced annotators. If there is no majority for an instance, its label would be "Unknown". We collect 1,119 unknown instances from SNLI dataset as our difficult instances, and collect 1,119 labeled instances from the instances of three classes (i.e., entailment, contradiction, and neutral) in equal proportion as our simple instances, obtaining a balanced binary classification (difficult or simple) dataset with 2,238 instances. We randomly sample 1,238 instances with balanced labels as training set and use the remaining 1,000 instances as test set.

\begin{figure}[t]
    \centering
    \includegraphics[width=.8\linewidth]{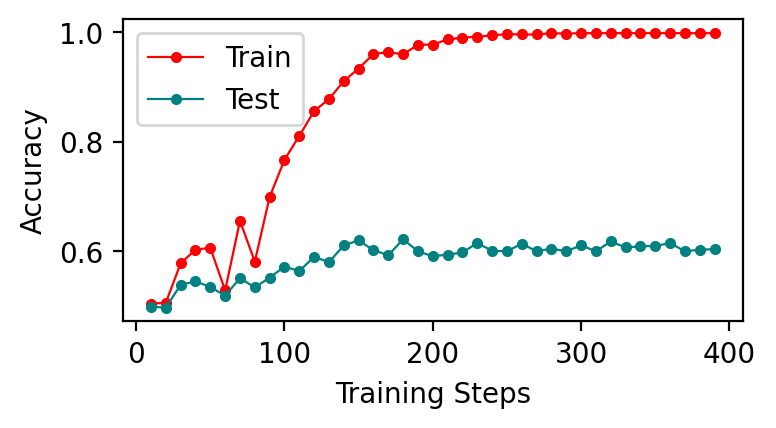}
    \caption{Training a BERT model to predict human-defined difficulty.}
    \label{fig:human_difficulty}
\end{figure}

\paragraph{Learning Human-defined Difficulty}
We then train a BERT model~\cite{Devlin2019BERT} with a linear classifier on the top on our constructed training set, and evaluate on the test set to see if it can predict whether an unseen instance is simple or difficult. As shown in Figure~\ref{fig:human_difficulty}, the BERT model that fits well on the training set can only achieve $\sim$60\% accuracy on the test set, demonstrating that neural models (even BERT) can not easily learn to estimate human-defined difficulty.

\subsection{Model-defined Difficulty}
However, model can have a different view of instance difficulty from human. For example, an instance can be defined as a difficult one if it can not be correctly predicted by a well-trained model. Thus, we also construct datasets to characterize model-defined difficulty for each instance, which is more realistic in the context of early exiting. In particular, we construct two datasets labeled with model-defined difficulty at sentence-level and token-level, respectively.


\begin{figure}[t]
    \centering
    \includegraphics[width=\linewidth]{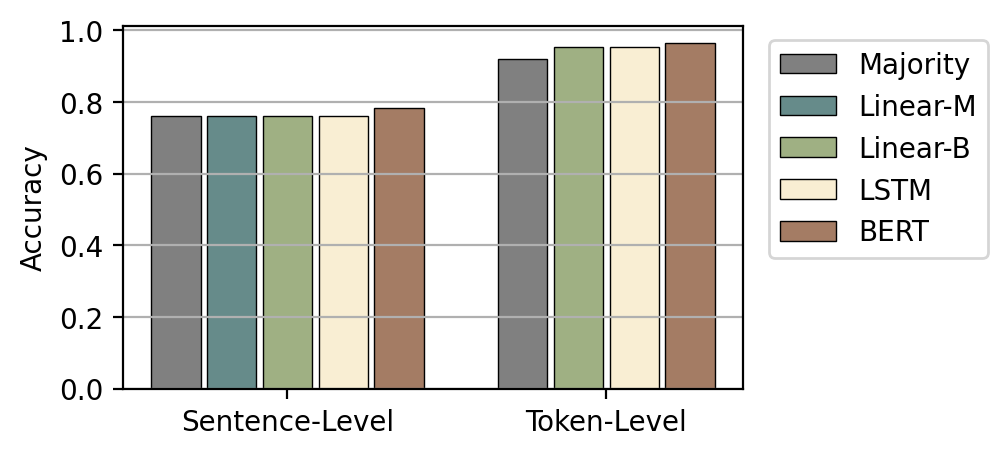}
\caption{The best accuracy achieved by different models on our constructed datasets for model-defined difficulty. The trained neural networks perform on par with the simple majority model.}
\label{fig:model_difficulty}
\end{figure}

\paragraph{Sentence-level Difficulty}
Estimating model-defined difficulty of a sentence (or sentence pairs) is helpful to language understanding tasks such as text classification and natural language inference~\cite{Xin2021BERxiT}. To obtain the sentence-level difficulty, we train a multi-exit BERT that is attached with an internal classifier at each layer on SNLI training set. Once the multi-exit BERT is trained, it can serve as an annotator to label each instance in the SNLI development set whether it can be correctly predicted by each internal classifier. In our experiments, we use BERT\textsubscript{BASE} that has 12 layers, and therefore for each instance in the SNLI development set we have 12 labels, each takes values of 0 or 1 to indicate whether or not the corresponding internal classifier correctly predict its label. By this, we label the 9,842 SNLI development instances to construct a multi-label classification dataset, from which we randomly sample 8,000 instances as training set and use the remaining 1,842 instances as test set.

\paragraph{Token-level Difficulty}
We also construct a dataset for estimating model-defined difficulty of each token, which can be used in language generation tasks~\cite{Elbayad2020Depth} and sequence labeling tasks~\cite{Li2020Accelerating}. Similarly, we train a multi-exit BERT on OntoNotes NER~\cite{Hovy2006Ontonotes} training set, and use it to annotate each token in the OntoNotes development instances whether it can be correctly predicted by each internal classifier. By this, we obtain a token-level multi-label classification dataset consisting of 13,900 instances, from which we randomly sample 10,000 instances to construct a training set and use the remaining 3,900 instances as test set.

\begin{table}[t]
\centering
\resizebox{.8\linewidth}{!}{
\begin{tabular}{lccc}
\toprule
\textbf{Model} & \textbf{Precision} & \textbf{Recall} & \textbf{F1 Score} \\ \midrule
\multicolumn{4}{c}{Sentence-Level Difficulty}                  \\
\midrule
Majority       & 60.5     & 36.7     & 45.7            \\
Linear-M       & 54.8     & 42.1     & 47.6            \\
Linear-B       & 52.9     & 45.3     & 48.8            \\
BiLSTM         & 54.5     & 45.2     & 49.4            \\
BERT           & 61.1     & 49.9     & 54.9            \\ \midrule
\multicolumn{4}{c}{Token-Level Difficulty}                     \\
\midrule
Majority\textsuperscript{*}       & -        & -        & -            \\
Linear-B       & 56.6     & 38.7     & 46.0         \\
BiLSTM         & 46.8     & 39.9     & 43.0         \\
BERT           & 65.6     & 44.6     & 53.1         \\ \bottomrule
\end{tabular}
}
\caption{Experimental results on our constructed model-defined difficulty datasets. We report micro-averaged precision, recall and F1 score over the negative label. *: The majority model for the token-level task would always predict positive class for all the labels, and therefore the F1 score is not applicable.}
\label{tab:model_difficulty}
\end{table}



\paragraph{Learning Model-defined Difficulty}
For each constructed model-defined difficulty dataset, we evaluate several models: (1) \textbf{Majority} model always predicts the majority class for each label, with class priors learned from the training data. (2) \textbf{Linear-M} is a multi-classification linear layer that takes as input the average pooled word embeddings and outputs the exiting layer. This model corresponds to the multinomial variants of \citet{Elbayad2020Depth}. Since the inputs of Linear-M is non-contextualized, we did not apply it to estimate token-level difficulty. (3) \textbf{Linear-B} is a binary classification linear layer that takes as input the hidden states at each BERT layer and outputs whether or not the instance (or token) is correctly predicted. This model corresponds to the geometric variants of \citet{Elbayad2020Depth} and the learn-to-exit module in BERxiT~\cite{Xin2021BERxiT}. (4) We also train and evaluate a bidirectional \textbf{LSTM} model~\cite{Hochreiter1997LSTM} with one layer and hidden size of 256. It takes as input the instance and outputs the exiting layer. (5) \textbf{BERT} model~\cite{Devlin2019BERT} is also considered for this task. For these models, except for Linear-B, we use the binary cross entropy loss to handle the multi-label classification. Since most development instances are correctly predicted, our constructed datasets are label-imbalanced. To alleviate this issue, we adopt over-sampling for classes with fewer instances.


Our experimental results are shown in Figure~\ref{fig:model_difficulty}, from which we find that: (1) For the task of estimating sentence-level difficulty, the shallow neural models perform as well as simple majority model. Only the BERT model can slightly outperform the majority model. (2) For token-level difficulty, these neural models perform slightly better than the majority model. The insignificant improvement over the majority model demonstrate that, the performance of the neural models mainly come from the learning of prior distribution of label instead of extracting difficulty-related features from instances. In the case of label imbalance, the accuracy can not well measure model performance. Besides, in the context of early exiting, we are more interested in cases that the model performs a false exit for an unsolved instance. Thus, we also report the precision, recall, and F1 score on the negative class. As shown in Table~\ref{tab:model_difficulty}, all the evaluated models perform poorly on recognizing the incorrectly predicted instances and tokens.


Though, it can not be concluded that the instance difficulty can not be learned since there are still a variety of machine learning models and training techniques that are under explored. Our preliminary experiments demonstrate that, at least, instance difficulty, whether human-defined or model-defined, is hard to learn for modern neural networks. In fact, our evaluated learn-to-exit models are upper baselines than that used in previous work because: (1) we also adopt more powerful deep models instead of simple linear models in previous methods~\cite{Elbayad2020Depth,Xin2021BERxiT}, and (2) Different from our method that trains learn-to-exit module on development set, previous methods jointly train their learn-to-exit module on the training set where few instances are incorrectly predicted, leading to more serious label imbalance. To facilitate future research, our constructed difficulty datasets will be publicly available.

\section{\abbr: Hash Early Exiting}
\subsection{What is Unnecessary and What Works?}
On the one hand, previous methods~\cite{Elbayad2020Depth,Xin2021BERxiT} that use learn-to-exit modules have achieved competitive results, which implies that something works in the learn-to-exit modules. On the other hand, our preliminary experiments show that instance difficulty is hard to be predicted in advance, which indicates that learning can be unnecessary to achieve a good performance.

To find what works, we formally describe the prediction of an early exiting model as $P(y|x)=\sum_{d\in \mathcal{D}}P(y|x,d)P(d|x)$, where $d$ is the difficulty (e.g., the exiting layer) for $x$. Note that in practice, $P(\mathcal{D}|x)$ is an one-hot distribution, so when $d$ is predicted, the exiting layer, i.e., the model architecture is determined. Therefore, the difficulty $d$ actually corresponds to an architecture.\footnote{Note that this formulation is similar to some differentiable Neural Architecture Search (NAS) and Mixture-of-Expert (MoE) works, which also encountered similar difficulties in learning architectures~\cite{Wang2021Rethinking,Roller2021Hash}.} Now given that the mapping from instance $x$ to its difficulty $d$, i.e., the best architecture, is hard to be learned, a natural idea to make $P(y|x)$ performs well is to keep $P(d|x)$ consistent: \textit{if a training instance $x_i$ is predicted to exit at layer $l$, then an inference instance $x_j$ that is similar with $x_i$ should exit at layer $l$, too.} By this, the activated architecture can well-handle the instance $x_j$ during inference because it is well-trained on similar instances such as $x_i$. Note that this consistency between training and inference can be easily satisfied by previous learn-to-exit modules due to the smoothness of neural models~\cite{Ziegel2003Elements}. Based on this hypothesis, we manage to remove the learning process and only stick to the consistency. In particular, we replace the neural learn-to-exit module $P(d|x)$ with a simple hash function.

\subsection{Method}
Without loss of generality, we first consider sequence classification tasks. A straightforward idea is to design a hash function to map semantically similar instances into the same bucket, and therefore the hash function should be some powerful sequence encoder such as Sentence-BERT~\cite{Reimers2019SBERT}, which is cumbersome in computation. In addition, a high-quality sequence encoder as a hash function usually maps instances with the same label into the same bucket (i.e. the same exiting layer), which makes the internal classifier at that layer suffer from label imbalance. Due to the difficulty of holding consistency at sentence-level, we rather propose to hold the consistency at token-level. By assigning each token into a fixed bucket, the token-level consistency between training and inference is easily satisfied.

\begin{figure}[t]
    \centering
    \includegraphics[width=\linewidth]{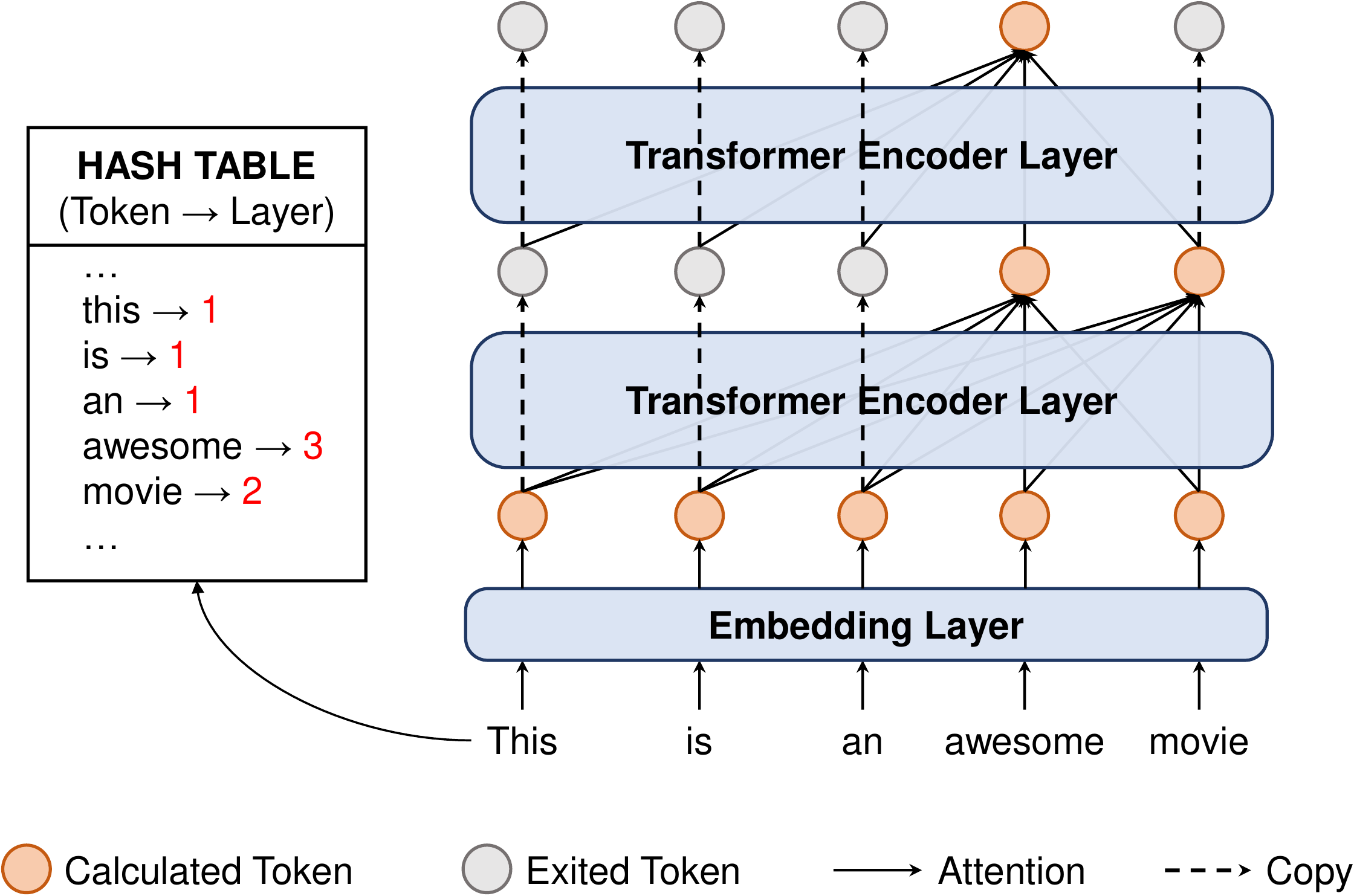}
    \caption{Overview of the Hash-based Early Exiting (\abbr). Tokens are assigned to fixed exiting layers using a hash function.}
    \label{fig:overview}
\end{figure}

An overview of our method is illustrated in Figure~\ref{fig:overview}. We adopt a simple and efficient hash function to map each token into a fixed bucket in advance, where each bucket corresponds to an exiting layer. We use pre-trained Transformers~\cite{Vaswani2017Attention} as our backbones. During model's forward pass, the representation of exited tokens will not be updated through self-attention, and its hidden states of the upper layers are directly copied from the hidden states of the exiting layer. By this token-level early exiting, the computation in self-attention and the following feed-forward network is reduced.

\subsection{Hash Functions}
\label{sec:hash}
To hold the token-level consistency between training and inference, \abbr~employs hash functions to compute in advance the exiting layer for each token. During training and inference, each token exits at a fixed layer according to the pre-computed hash lookup table. The hash functions can take a variety of forms. Here we consider several hash functions as possible alternatives.

\paragraph{Random Hash} Random hash is a lower baseline, wherein we assign each token to a fixed, random exiting layer at initialization. To examine our hypothesis, we also consider to use two different random hash functions for training and inference respectively, in which case the consistency does not hold. We denote these two random hash functions as \textbf{Rand-cons} and \textbf{Rand-incons}.

\paragraph{Frequency Hash} To achieve higher speed-up, a natural way is to assign frequent tokens to lower layers to exit. Intuitively, frequent tokens are usually well-trained during pre-training and therefore do not require too much refinement by looking at their contexts. Thus we can design a hash function that assigns tokens into exiting layers by frequency. In particular, the tokens are sorted by frequency and then divided equally into $B$ buckets.

\paragraph{MI Hash} Further, we also consider a task-specific hash function that is based on the mutual information (MI) between each token and the corresponding label, which, as an instance of \abbr, is also adopted in \citet{Liu2021Faster}. Tokens are sorted by their MI values between the task label, and then divided equally into $B$ buckets. Tokens with higher MI values are assigned to lower layers.

\paragraph{Clustered Hash} It is also intuitive that similar tokens should be assigned to the same layer to exit, and therefore we also experiment with a clustered hash function. The clusters are obtained by performing k-means clustering using token embeddings from BERT\textsubscript{BASE} embedding layer. The clustered tokens are then sorted by norm, which often relates to token frequency~\cite{Schakel2015Measuring} and difficulty~\cite{Liu2020Norm}. The clustered tokens with small average norms are assigned to lower layers.

\section{Experiments}

\begin{table*}[t]
\centering
\resizebox{.95\linewidth}{!}{
\begin{tabular}{lccccccr}
\toprule
\multirow{2}{*}{\textbf{Models}}               & \textbf{SST-2}       & \textbf{IMDb}        & \textbf{SNLI}        & \textbf{SciTail}     & \textbf{MRPC}        & \textbf{STS-B}       & \textbf{ELUE} \\
& (8.5k) & (20.0k) & (549.4k) & (23.6k) & (3.7k) & (5.7k) & \textbf{Score} \\ \midrule
\multicolumn{8}{l}{\textit{Pre-Trained Language Models}}                                                                       \\
BERT-3L              & 79.3 \small{(4.0$\times$)} & 88.4 \small{(4.0$\times$)} & 87.1 \small{(4.0$\times$)} & 84.3 \small{(4.0$\times$)} & 76.0 \small{(4.0$\times$)} & 75.8 \small{(4.0$\times$)} & -3.70       \\
ALBERT-3L            & 82.4 \small{(3.6$\times$)} & 90.7 \small{(3.9$\times$)} & 87.8 \small{(3.7$\times$)} & 87.5 \small{(3.9$\times$)} & 80.0 \small{(3.6$\times$)} & 79.1 \small{(3.9$\times$)} & -1.59       \\
RoBERTa-3L           & 81.8 \small{(4.1$\times$)} & 90.7 \small{(4.2$\times$)} & 88.0 \small{(3.8$\times$)} & 84.9 \small{(3.9$\times$)} & 75.6 \small{(3.9$\times$)} & 67.5 \small{(3.9$\times$)} & -2.17       \\
ElasticBERT-3L       & 84.1 \small{(4.0$\times$)} & 91.8 \small{(4.0$\times$)} & 89.3 \small{(4.0$\times$)} & 91.9 \small{(4.0$\times$)} & 83.1 \small{(4.0$\times$)} & 83.5 \small{(4.0$\times$)} & 0.00        \\ \midrule
\multicolumn{8}{l}{\textit{Static Models}}                                                                                     \\
DistilBERT           & 84.8 \small{(2.0$\times$)} & 92.0 \small{(2.0$\times$)} & 89.2 \small{(2.0$\times$)} & 89.7 \small{(2.0$\times$)} & 83.8 \small{(2.0$\times$)}  & 81.7 \small{(2.0$\times$)} & -2.55        \\
TinyBERT             & \underline{85.3} \small{(2.0$\times$)} & 89.0 \small{(2.0$\times$)} & 89.3 \small{(2.0$\times$)} & 90.0 \small{(2.0$\times$)} & \textbf{84.7} \small{(2.0$\times$)}  & \textbf{85.0} \small{(2.0$\times$)} & -2.20       \\
HeadPrune            & 84.8 \small{(1.3$\times$)}        & 84.7 \small{(1.5$\times$)}        & 87.8 \small{(1.5$\times$)}       & 88.3 \small{(1.5$\times$)}        & 77.8 \small{(1.5$\times$)}        & 74.8 \small{(1.5$\times$)}        & -6.85            \\
BERT-of-Theseus      & 84.4 \small{(2.0$\times$)}        & 90.7 \small{(2.0$\times$)}        & \underline{89.4} \small{(2.0$\times$)}        & \underline{92.1} \small{(2.0$\times$)}        & 82.4 \small{(2.0$\times$)}        & \textbf{85.0} \small{(2.0$\times$)}        & -2.55            \\ \midrule
\multicolumn{8}{l}{\textit{Dynamic Models}}                                                                                    \\
DeeBERT              & 78.9 \small{(3.4$\times$)} & 79.5 \small{(4.1$\times$)} & 48.1 \small{(3.6$\times$)} & 71.9 \small{(3.4$\times$)} & 79.1 \small{(3.5$\times$)} & -           & -          \\
FastBERT             & 82.7 \small{(3.7$\times$)} & \textbf{92.5} \small{(3.5$\times$)} & 88.8 \small{(3.5$\times$)} & 89.0 \small{(3.6$\times$)} & 80.3 \small{(4.2$\times$)} & -           & -          \\
PABEE                & 83.1 \small{(2.9$\times$)} & 91.6 \small{(3.4$\times$)} & 88.7 \small{(3.1$\times$)} & 90.7 \small{(3.3$\times$)} & 75.2 \small{(3.5$\times$)} & 80.1 \small{(3.2$\times$)} & -1.31      \\
CascadeBERT          & 82.4 \small{(3.8$\times$)}            & 91.8 \small{(3.7$\times$)}            & 89.0 \small{(3.6$\times$)}            & 91.7 \small{(3.8$\times$)}            &  78.8 \small{(3.8$\times$)}           & -            & -           \\
BERxiT w/ BERT       & 71.8 \small{(2.2$\times$)} & 85.0 \small{(2.8$\times$)}   & 88.4 \small{(3.6$\times$)} & 80.3 \small{(3.4$\times$)} & 74.9 \small{(4.0$\times$)} & 57.8 \small{(4.0$\times$)} & -6.12      \\
BERxiT w/ ElasticBERT& 72.6 \small{(4.4$\times$)} & 91.2 \small{(4.0$\times$)}   & 84.7 \small{(3.9$\times$)} & 91.0 \small{(4.0$\times$)} & 78.6 \small{(4.3$\times$)} & 81.5 \small{(4.0$\times$)} & -3.90      \\
\midrule
\multicolumn{8}{l}{\textit{Ours}} \\
\abbr                & \textbf{85.5} \small{\textbf{(4.8}$\times$)} & \underline{92.4} \small{(\textbf{6.2}$\times$)} & \textbf{89.6} \small{(\textbf{4.4}$\times$)} & \textbf{92.3} \small{(\textbf{5.1}$\times$)} & \underline{84.0} \small{(\textbf{4.8}$\times$)} & 84.3 \small{(\textbf{4.6}$\times$)} & \textbf{1.20}           \\ \bottomrule

\end{tabular}
}
\caption{Main results on the ELUE benchmark~\cite{Liu2021ELUE}. We report for each model on each task the performance and the corresponding speedup ratio, which is calculated as the FLOPs reduction relative to BERT\textsubscript{BASE}. For MRPC, we report the mean of
accuracy and F1. For STS-B, we report Pearson and Spearman correlation. For all other tasks we report accuracy. "-" indicates that the method is not applicable on that task.}
\label{tab:ELUE}
\end{table*}

\subsection{Tasks and Datasets}
Since \abbr~requires no supervision, it can be applied to a variety of tasks and architectures. In our work, we conduct experiments on natural language understanding tasks including sentiment analysis, natural language inference, similarity regression, and a language generation task, text summarization. Statistics of our used datasets are shown in Appendix~\ref{sec:dataset_stat}.

\paragraph{Understanding Tasks}
For the convenience of comparison with other efficient models, we evaluate our proposed \abbr~on the ELUE benchmark~\cite{Liu2021ELUE}, which is comprised of SST-2~\cite{Socher2013SST}, IMDb~\cite{Maas2011IMDb}, SNLI~\cite{Bowman2015SNLI}, SciTail~\cite{Khot2018SciTail}, MRPC~\cite{Dolan2005MRPC}, and STS-B~\cite{Cer2017SemEval2017}). Note that STS-B is a regression task.

\paragraph{Generation Tasks}
For language generation, we evaluate \abbr~on two English summarization datasets, CNN/DailyMail~\cite{Hermann2015CNNDM} and Reddit~\cite{Kim2019Reddit}, and two Chinese summarization datasets: TTNews~\cite{hua2017overview} and CSL~\cite{Xu2020CLUE}.

\subsection{Experimental Setup}
\paragraph{Baselines}
We compare \abbr~with the following competitive baseline models: \textbf{(1) Pre-Trained Language Models}. We directly fine-tune the first layers of pre-trained language models including BERT~\cite{Devlin2019BERT}, ALBERT~\cite{Lan2020ALBERT}, RoBERTa~\cite{Liu2019roberta}, and ElasticBERT~\cite{Liu2021ELUE} with a MLP classifier on the top. \textbf{(2) Static Models}. We compare with several static approaches to accelerate language model inference, including DistilBERT~\cite{Sanh2019DistilBERT}, TinyBERT~\cite{Jiao2020TinyBERT}, HeadPrune~\cite{Michel2019HeadPrune}, and BERT-of-Theseus~\cite{Xu2020BERTTheseus}. \textbf{(3) Dynamic models}. We compare with DeeBERT~\cite{Xin2020DeeBERT}, FastBERT~\cite{Liu2020FastBERT}, PABEE~\cite{Zhou2020PABEE}, BERxiT~\cite{Xin2021BERxiT}, and CascadeBERT~\cite{Li2021cascadebert}.

\paragraph{Training}
For most NLU experiments we adopt the ElasticBERT\textsubscript{BASE} model~\cite{Liu2021ELUE} as our backbone model, which is a pre-trained multi-exit Transformer encoder. For small datasets (i.e., SST-2, MRPC, and STS-B) we report the mean performance and the standard deviation (in Table~\ref{tab:hash} and \ref{tab:bucket}) over 5 runs with different random seeds. For text summarization datasets we adopt BART\textsubscript{BASE}~\cite{Lewis2020BART} and CPT\textsubscript{BASE}~\cite{Shao2021CPT} as our backbone models and use the frequency hash to assign tokens to the encoder layers. All of the experiments are conducted on GeForce RTX 3090 GPUs. More experimental details are given in Appendix~\ref{sec:detail}.



\subsection{Results and Analysis}
\label{sec:results}
\begin{table}[t]
\centering
\resizebox{\linewidth}{!}{
\begin{tabular}{l|c|ccc}
\toprule
\textbf{Hash} & \textbf{Speed} & \textbf{SST-2} & \textbf{SNLI} & \textbf{MRPC}  \\
\textbf{Functions}    & \textbf{-up}            & (8.5k)         & (549.4k)      & (3.7k) \\ \midrule
\multicolumn{5}{c}{Backbone: ElasticBERT-6L} \\ \midrule
Rand-incons                            & 3.0$\times$           & 85.5 \small{($\pm$0.53)}           & 89.7          & 85.0 \small{($\pm$0.22)} \\
Rand-cons                              & 3.0$\times$           & 85.7 \small{($\pm$0.45)}           & 90.1          & 86.3 \small{($\pm$0.67)} \\
Frequency                              & 4.9$\times$           & 85.5 \small{($\pm$0.41)}           & 89.6          & 84.0 \small{($\pm$0.27)} \\
MI                                     & 3.3$\times$           & 85.5 \small{($\pm$0.49)}           & 90.0          & 86.0 \small{($\pm$0.23)} \\
Clustered                              & 3.0$\times$           & 85.7 \small{($\pm$0.50)}           & 90.2          & 86.3 \small{($\pm$0.47)} \\ \midrule
\multicolumn{5}{c}{Backbone: ElasticBERT-12L} \\ \midrule
Rand-incons                            & 1.6$\times$           & 85.7 \small{($\pm$0.38)}           & 89.6          & 86.6 \small{($\pm$0.45)} \\
Rand-cons                              & 1.5$\times$           & 86.5 \small{($\pm$0.37)}           & 90.2          & 87.4 \small{($\pm$0.34)} \\
Frequency                              & 2.8$\times$           & 85.6 \small{($\pm$0.37)}           & 89.8          & 84.4 \small{($\pm$0.17)} \\
MI                                     & 1.8$\times$           & 86.6 \small{($\pm$0.17)}           & 90.1          & 87.2 \small{($\pm$0.66)}\\
Clustered                              & 1.5$\times$           & 87.0 \small{($\pm$0.54)}           & 90.1          & 87.3 \small{($\pm$0.48)} \\ \bottomrule
\end{tabular}
}
\caption{Comparison of different hash functions. The speed-up ratios are calculated by FLOPs reduction relative to BERT\textsubscript{BASE} and averaged over the three tasks. The ELUE score is averaged over the three tasks. For small datasets, i.e., SST-2 and MRPC, we report the mean and standard deviation over five runs.}
\label{tab:hash}
\end{table}

\begin{figure}[t]
    \centering
    \includegraphics[width=.8\linewidth]{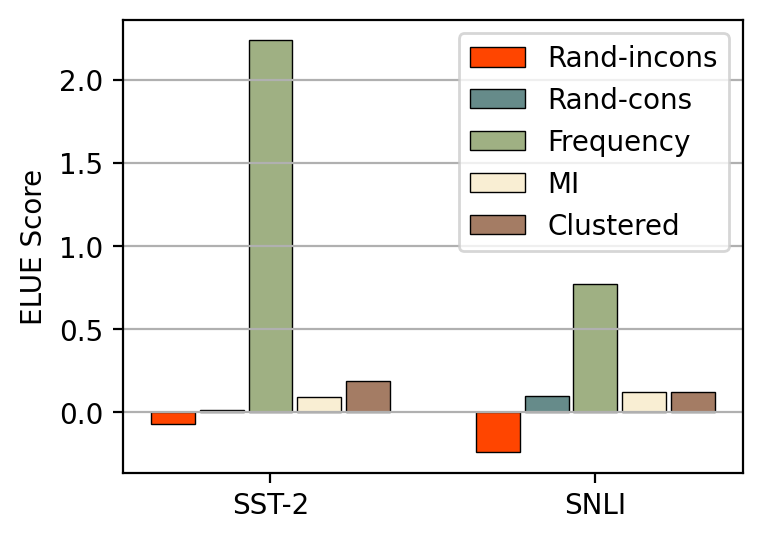}
    \caption{Comparison of the ELUE scores achieved by \abbr~with different hash functions.\vspace{-0.2cm}}
    \label{fig:hash_func}
\end{figure}

\begin{figure}[t]
    \centering
    \begin{subfigure}{.49\linewidth}
    \centering
    \includegraphics[height=1.2\linewidth]{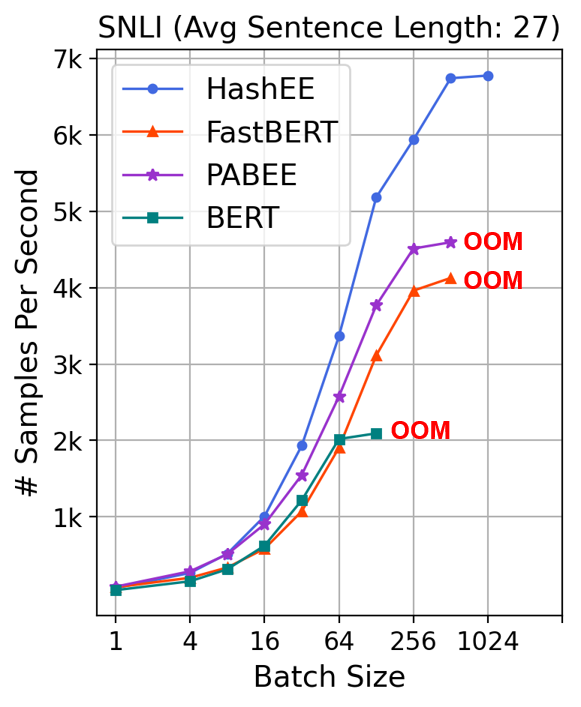}
    \end{subfigure}
    \begin{subfigure}{.49\linewidth}
    \centering
    \includegraphics[height=1.2\linewidth]{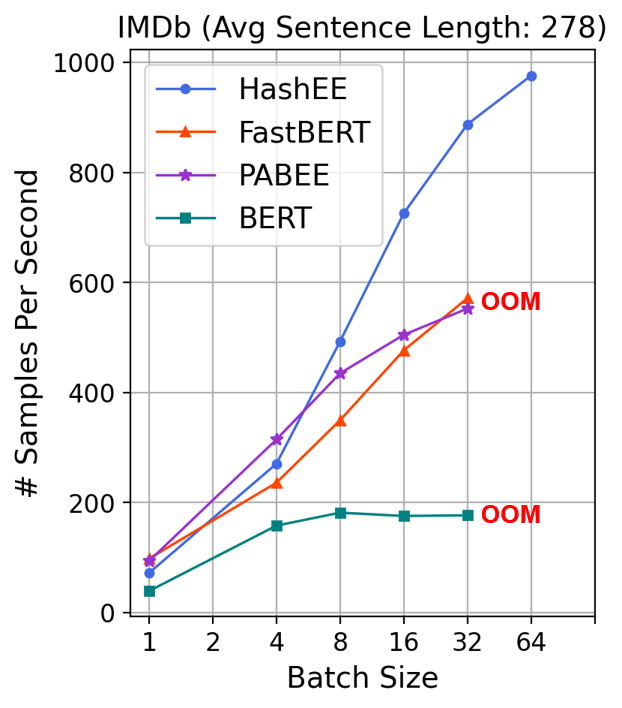}
    \end{subfigure}
    \caption{Comparison of actual inference time.\vspace{-0.5cm}}
    \label{fig:inference}
\end{figure}

\begin{figure}[t]
    \centering
    \includegraphics[width=\linewidth]{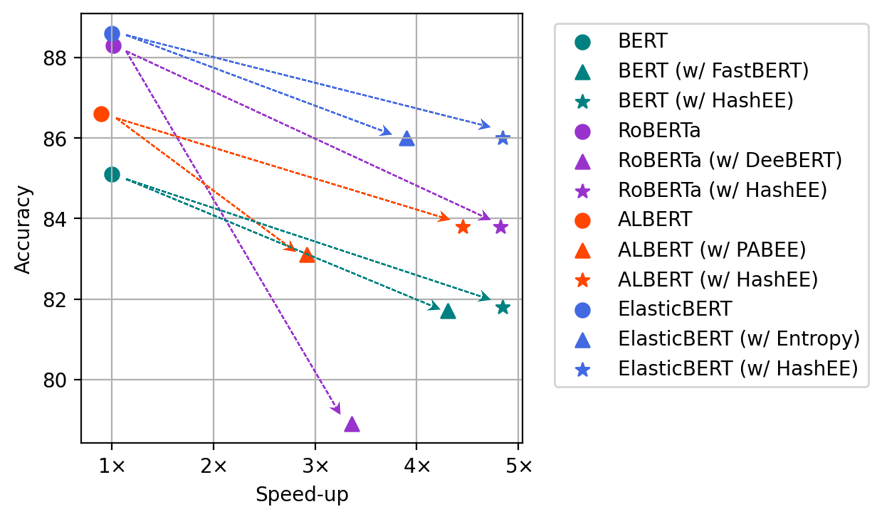}
    \caption{Comparison of different backbone models on ELUE SST-2 dataset.\vspace{-0.5cm}}
    \label{fig:backbone}
\end{figure}


\begin{table*}[t]
\resizebox{\linewidth}{!}{
\begin{tabular}{l|ccc|cc|cc}
\toprule
\multirow{2}{*}{\textbf{Model}} & \multicolumn{3}{c|}{\textbf{Speed-up}} & \multicolumn{2}{c|}{\textbf{English}} & \multicolumn{2}{c}{\textbf{Chinese}} \\
                                & Enc.        & Dec.       & Total       & \textbf{Reddit}   & \textbf{CNN/DM}   & \textbf{CSL}       & \textbf{TTNews}  \\ \midrule
BART                            & 1.0$\times$        & 1.0$\times$       & 1.0$\times$        & 29.71/9.91/23.43  & 44.16/21.28/40.90 &  64.49/52.48/61.81                  &      53.84/38.09/49.85           \\
DAT                             & 1.0$\times$        &   0.5$\times$         &     0.8$\times$        & 27.02/\textbf{8.89}/\textbf{22.68}  &      40.30/17.77/37.53             & -                   & -                \\
BART-6L                         & 2.0$\times$        & \textbf{1.4}$\times$       & \textbf{1.8}$\times$        &   26.22/6.82/21.05  & 40.02/16.60/36.82         &                   & -                                  \\
\abbr~w/ BART                   & \textbf{3.3}$\times$        & 1.0$\times$       & \textbf{1.8}$\times$        &     \textbf{28.77}/8.52/21.97             &    \textbf{41.04}/\textbf{18.41}/\textbf{37.65}               & -                   & -                \\
\midrule
CPT                             & 1.0$\times$        & 1.0$\times$       & 1.0$\times$           & -                 & -                 & 65.49/53.82/62.96  & 53.48/37.59/49.82                \\
CPT-6L                          & 2.0$\times$        & \textbf{1.2}$\times$       & 1.9$\times$           & -                 & -                 & 52.29/39.35/50.06  & 50.89/33.75/45.42       \\
\abbr~w/ CPT                    & \textbf{2.3}$\times$         & 1.0$\times$       &  \textbf{2.2}$\times$           & -                  & -                  & \textbf{62.42}/\textbf{49.96}/\textbf{59.15}                   & \textbf{52.67}/\textbf{35.31}/\textbf{46.97}                \\
\bottomrule
\end{tabular}
}
\caption{Experimental results on two English and two Chinese summarization datasets. We report ROUGE-1, ROUGE-2, and ROUGE-L for each dataset. The speedup ratios for English and Chinese models are calculated by the FLOPs reduction relative to BART\textsubscript{BASE} and CPT\textsubscript{BASE}, respectively, and averaged over the performed datasets. Here we re-implement the confidence thresholding variant of DAT~\cite{Elbayad2020Depth}.\vspace{-0.4cm}}
\label{tab:summ}
\end{table*}


\paragraph{Results on ELUE}
We first show our main comparison results on ELUE test sets in Table~\ref{tab:ELUE}. Using the frequency hash that assigns tokens to the first 6 layers of ElasticBERT\textsubscript{BASE}, \abbr~can outperform most considered baselines with fewer FLOPs. To fairly compare with baselines of various speedup ratios, we also report the ELUE score~\cite{Liu2021ELUE}, which is a two-dimensional (performance and FLOPs) metric for efficient NLP models, measuring how much a model oversteps ElasticBERT. Table~\ref{tab:ELUE} shows that \abbr~achieves a new state-of-the-art ELUE score. To fairly compare with the learn-to-exit baseline we also implement BERxiT~\cite{Xin2021BERxiT} with ElasticBERT\textsubscript{BASE}.

\paragraph{Comparison of Different Hash Functions}
We then evaluate \abbr~with different hash functions detailed in Section~\ref{sec:hash}. For all these hash functions, we assign tokens to the 6 and 12 layers of ElasticBERT-6L and ElasticBERT-12L, respectively. Experimental results on SST-2, SNLI, and MRPC are given in Table~\ref{tab:hash}. Among the hash functions, the frequency hash achieves the highest speedup while maintaining a considerable performance. With the backbone of ElasticBERT-12L, these hash functions, except for the frequency hash, cannot achieve considerable speedup. Besides, we find that ElasticBERT-12L did not significantly outperform ElasticBERT-6L with \abbr. We conjecture that higher layers are not good at querying information from hidden states of tokens that exit too early. In this work, we are more interested in the case of high acceleration ratio, so we adopt ElasticBERT-6L as our main backbone. To make a more intuitive comparison of these hash functions with different speedup ratios, we also show the ELUE scores on SST-2 and SNLI with ElasticBERT-6L as backbone in Figure~\ref{fig:hash_func} . We find that the frequency hash outperforms other hash functions by a large margin, and therefore in the following experiments we mainly use the frequency hash. Besides, only the Rand-incons hash obtains negative ELUE score, demonstrating the benefit of maintaining consistency between training and inference.

\paragraph{Comparison of Actual Inference Time}
Because most of the operations in the Transformer architecture are well optimized by modern deep learning frameworks and parallel processing hardwares such as GPU and TPU, FLOPs may not precisely reflect the actual inference time. To that end, here we also evaluate actual inference time on a single GeForce RTX 3090 GPU. Note that the speedup ratio of previous early exiting methods are usually tested on a per-instance basis, i.e. the batch size is set to 1. However, batch inference is often more favorable in both offline scenarios and low-latency scenarios~\cite{Zhang2019MArk}. Here we compare \abbr~with two baselines that have similar performance, i.e., FastBERT and PABEE. Our experiments are conducted on two datasets with very different average sentence length, i.e., SNLI and IMDb. Results are given in Table~\ref{tab:max_speed} and Figure~\ref{fig:inference}. We find \abbr~has an advantage in processing speed when the batch size exceeds 8. Besides, \abbr~can perform larger batch inference due to its memory-efficiency.

\begin{table}[h]
\resizebox{\linewidth}{!}{
\begin{tabular}{l|ccc|ccc}
\toprule
         & \multicolumn{3}{c|}{\textbf{SNLI (Avg Len: 27)}} & \multicolumn{3}{c}{\textbf{IMDb (Avg Len: 278)}} \\
         & Acc      & Speed-up    & \# samples/sec    & Acc      & Speed-up    & \# samples/sec    \\ \midrule
BERT     & 90.4     & 1.0$\times$     & 2093 \small{(1.0$\times$)}       & 93.0     & 1.0$\times$     & 177 \small{(1.0$\times$)}        \\
FastBERT & 88.8     & 3.5$\times$     & 4128 \small{(2.0$\times$)}       & 92.5     & 3.5$\times$     & 553 \small{(3.1$\times$)}        \\
PABEE    & 88.7     & 3.1$\times$     & 4596 \small{(2.2$\times$)}       & 91.6     & 3.4$\times$     & 571 \small{(3.2$\times$)}        \\
HashEE   & 89.6     & 4.4$\times$     & 6779 \small{(3.2$\times$)}       & 92.4     & 6.2$\times$     & 976 \small{(5.5$\times$)}        \\ \bottomrule
\end{tabular}
}
\caption{Maximal number of processing samples per second on a single RTX 3090 GPU.}
\label{tab:max_speed}
\end{table}

\paragraph{Comparison of Different Backbones}
To evaluate the versatility of \abbr, we also conduct experiments with other backbone models, i.e., BERT, ALBERT, and RoBERTa. As shown in Figure~\ref{fig:backbone}, \abbr~outperforms other baselines with the same backbone.

\paragraph{Accelerating Seq2Seq Models}
Since \abbr~requires no supervision, it can also be applied to seq2seq models for generation tasks. We first evaluate \abbr~with BART\textsubscript{BASE} as our backbone on two English summarization tasks. As shown in Table~\ref{tab:summ}, \abbr~can achieve significant speedup for BART encoder while maintaining considerable ROUGE scores. Besides, we find that previous early exiting methods that measure the uncertainty of internal outputs would rather slow down decoder inference due to the heavy computation of prediction over large vocabulary. In addition, to further explore the speedup potential of \abbr, we also experiment with CPT~\cite{Shao2021CPT}, which has a deep encoder and a shallow decoder. Results on CSL and TTNews depict that \abbr~can achieve 2.2$\times$ speedup relative to CPT while maintaining 97\% ROUGE-1. We also report results of the 6-layer versions of BART (with 3 encoder layers and 3 decoder layers) and CPT (with 5 encoder layers and 1 decoder layer).


\section{Related Work}
Large-scale pre-trained language models (PLMs) have achieved great success in recent years. Despite their power, the inference is time-consuming, which hinders their deployment in low-latency scenarios. To accelerate PLM inference, there are currently two streams of work: (1) Compressing a cumbersome PLM through knowledge distillation~\cite{Sanh2019DistilBERT,Sun2019PKD,Jiao2020TinyBERT}, model pruning~\cite{Gordon2020Compressing,Michel2019HeadPrune}, quantization~\cite{Shen2020QBERT}, module replacing~\cite{Xu2020BERTTheseus}, etc. (2) Selectively activating parts of the model conditioned on the input, such as Universal Transformer~\cite{Dehghani2019Universal}, FastBERT~\cite{Liu2020FastBERT}, DeeBERT~\cite{Xin2020DeeBERT}, PABEE~\cite{Zhou2020PABEE}, LeeBERT~\cite{Zhu2021Leebert}, CascadeBERT~\cite{Li2021cascadebert}, ElasticBERT~\cite{Liu2021ELUE} and other similar methods~\cite{Elbayad2020Depth,Schwartz2020Right,Liao2021Global,Xin2021BERxiT,Sun2021Early}. Different from these methods, our proposed \abbr~requires no internal classifiers (which imply extra parameters) and supervision, and therefore can be widely used in a variety of tasks and model architectures.

\section{Conclusion}
We first empirically study the learnability of instance difficulty, which is a crucial problem in early exiting. Based on the observation that modern neural models perform poorly on estimating instance difficulty, we propose a hash-based early exiting approach, named \abbr, that removes the learning process and only sticks to the consistency between training and inference. Our experiments on classification, regression, and generation tasks show that \abbr~can achieve state-of-the-art performance with fewer computation and inference time.


\section*{Acknowledgements}
This work was supported by the National Key Research and Development Program of China (No. 2020AAA0106702) and National Natural Science Foundation of China (No. 62022027).

\section*{Ethical Considerations}
Our proposed \abbr~aims to accelerate the inference of large-scale pre-trained languagde modles, and therefore helps reduce computation cost and carbon emission. Since our method simply reduces computation of some input tokens, it would not introduce significant new ethical concerns. All the datasets used in our experiments are from previously published papers, and to our knowledge, do not have any attached privacy or ethical issues. Nevertheless, further efforts should be made to study the biases encoded in the pre-trained language models and widely used datasets.

\bibliography{custom}
\bibliographystyle{acl_natbib}

\newpage
\appendix
\section{Appendix}
\subsection{Dataset Statistics}
\label{sec:dataset_stat}
Here we list the statistics of our used language understanding and generation datasets in Table~\ref{tab:dataset} and Table~\ref{tab:summ_data}.

\begin{table}[h]
\centering
\resizebox{\linewidth}{!}{
\begin{tabular}{llrrr}
\toprule
\textbf{Tasks}                                                                          & \textbf{Datasets} & \textbf{|Train|} & \textbf{|Dev|} & \textbf{|Test|} \\ \midrule
\multirow{2}{*}{\begin{tabular}[c]{@{}l@{}}Sentiment\\ Analysis\end{tabular}}           & SST-2             & 8,544            & 1,101          & 2,208           \\
                                                                                        & IMDb              & 20,000           & 5,000          & 25,000          \\ \midrule
\multirow{2}{*}{\begin{tabular}[c]{@{}l@{}}Natural Language\\ Inference\end{tabular}} & SNLI              & 549,367          & 9,842         & 9,824          \\
                                                                                        & SciTail           & 23,596           & 1,304          & 2,126           \\ \midrule
\multirow{2}{*}{\begin{tabular}[c]{@{}l@{}}Similarity and\\ Paraphrase\end{tabular}}  & MRPC              & 3,668            & 408            & 1,725           \\
                                                                                        & STS-B             & 5,749            & 1,500          & 1,379           \\ \bottomrule
\end{tabular}
}
\caption{Statistics of our used language understanding datasets.}
\label{tab:dataset}
\end{table}

\begin{table}[h]
\resizebox{\linewidth}{!}{
\begin{tabular}{llccc}
\toprule
\multirow{2}{*}{\textbf{Datasets}} & \multirow{2}{*}{\textbf{Source}} & \multicolumn{3}{c}{\textbf{\# Pairs}} \\
                                   &                                  & Train      & Dev       & Test      \\ \midrule
Reddit                             & Social Media                     & 41,675     & 645       & 645       \\
CNN/DM                             & News                             & 287,084     & 13,367     & 11,489     \\
TTNews                             & News                             & 50,000      & -          & 2,000      \\
CSL                                & Academic                         & 20,000      & 3,000      & 3,000      \\ \bottomrule
\end{tabular}
}
\caption{Statistics of our used text summarization datasets.}
\label{tab:summ_data}
\end{table}

\subsection{Experimental Details}
\label{sec:detail}
For small datasets in ELUE, i.e. SST-2, MRPC, and STS-B, we conduct grid search over batch sizes of \{16, 32\}, learning rates of \{2e-5, 3e-5, 5e-5\}, number of epochs of \{3, 4, 5\}, warmup step ratios of \{0.1, 0.01\}, and weight decays of \{0.1, 0.01\} with an AdamW~\cite{Ilya2019AdamW} optimizer. We select the hyperparameters that achieved the best performance on the development sets, and perform 5 runs with different random seeds to obtain the mean performance and standard deviation. For SNLI, SciTail, and IMDb, we use the same hyperparameters. The best-performed hyperparameters in our language understanding experiments are given in Table~\ref{tab:hyper}.

\begin{table}[h]
\centering
\resizebox{.9\linewidth}{!}{
\begin{tabular}{lccccc}
\toprule
\textbf{Tasks}   & \textbf{LR}   & \textbf{BSZ} & \textbf{Epoch} & \textbf{WSR} & \textbf{WD}   \\
\midrule
SST-2   & 5e-5 & 16  & 3     & 0.1 & 0.1  \\
IMDb    & 5e-5 & 32  & 3     & 0.1 & 0.01 \\
SNLI    & 5e-5 & 32  & 3     & 0.1 & 0.01 \\
SciTail & 5e-5 & 32  & 3     & 0.1 & 0.01 \\
MRPC    & 5e-5 & 32  & 4     & 0.1 & 0.01 \\
STS-B   & 5e-5 & 16  & 5     & 0   & 0.1 \\ \bottomrule
\end{tabular}
}
\caption{Best-performed hyperparameters on ELUE tasks. LR: Learning Rate. BSZ: Batch Size. WSR: Warmup Step Ratio. WD: Weight Decay.}
\label{tab:hyper}
\end{table}

For English summarization tasks, i.e., CNN/DailyMail and Reddit, we use the same hyperparameters as BART. For Chinese summarization tasks, i.e., TTNews and CSL, we use the same hyperparameters as CPT.

\subsection{Additional Experimental Results}
\label{sec:additional_res}
In previous experiments we assign tokens to the same number of buckets as the number of layers. Here we also explore other configurations. For each configuration, we assign tokens to $B$ buckets, corresponding to exiting layers $\{1+12b/B\}_{b=0}^{B-1}$. For instance, if we have 12 layers and 3 buckets, the 3 buckets correspond to the \{1, 5, 9\} layers. Overall results are given in Table~\ref{tab:bucket}, where we show results of 8 configurations with the frequency hash. Similar with Table~\ref{tab:hash}, we find that 6-layer models perform well while achieving higher acceleration ratios. In addition, the number of buckets has no significant effect on acceleration ratio. Configurations that the number of layers equals to the number of buckets perform slightly better than other configurations.

\begin{table}[t]
\centering
\resizebox{\linewidth}{!}{
\begin{tabular}{lc|c|ccc}
\toprule
\multirow{2}{*}{\textbf{\# L}} & \multirow{2}{*}{\textbf{\# B}} & \textbf{Speed} & \textbf{SST-2} & \textbf{SNLI} & \textbf{MRPC} \\
                                    &                                      & \textbf{-up}   & (8.5k)         & (549.4k)      & (3.7k) \\ \midrule
\multirow{5}{*}{12}                 & 12                                   & 2.8$\times$           & 85.6 \small{($\pm$0.37)}    & 89.8           & 84.4 \small{($\pm$0.17)}  \\
                                    & 6                                    & 2.9$\times$           & 84.9 \small{($\pm$0.69)}    & 89.7           & 83.7 \small{($\pm$0.26)} \\
                                    & 4                                    & 3.0$\times$           & 85.2 \small{($\pm$0.43)}    & 89.6           & 83.7 \small{($\pm$0.15)} \\
                                    & 3                                    & 3.0$\times$           & 85.3 \small{($\pm$0.37)}    & 89.7           & 82.9 \small{($\pm$0.29)} \\
                                    & 2                                    & 3.1$\times$           & 85.2 \small{($\pm$0.19)}    & 89.7           & 82.8 \small{($\pm$0.40)} \\ \midrule
\multirow{3}{*}{6}                  & 6                                    & 4.9$\times$           & 85.5 \small{($\pm$0.41)}    & 89.6          & 84.0 \small{($\pm$0.27)} \\
                                    & 3                                    & 5.0$\times$           & 85.2 \small{($\pm$0.42)}    & 89.5          & 83.5 \small{($\pm$0.54)} \\
                                    & 2                                    & 5.1$\times$           & 85.4 \small{($\pm$0.33)}    & 89.6          & 83.6 \small{($\pm$0.19)} \\ \bottomrule
\end{tabular}
}
\caption{Comparison of different numbers of model layers and buckets with frequency hash function. "\# L" and "\# B" mean number of layers and number of buckets. For small datasets, i.e., SST-2 and MRPC, we report the mean and standard deviation over five runs with different random seeds.}
\label{tab:bucket}
\end{table}

\subsection{Details on FLOPs Calculation}
Here we take a closer look at the \abbr~model forward process, and see which FLOPs are saved during inference.

Given the hidden states at layer $l$ as $\mathbf{H}^l\in\mathbb{R}^{n\times d}$ and the hidden states of remaining tokens are denoted as $\mathbf{h}^l\in\mathbb{R}^{m\times d}$, where $n$ is the original sequence length and $m$ is the number of remaining tokens at layer $l$, the calculation of one Transformer encoder layer with \abbr~ can be formally described as
\begin{align}
	\mathbf{q}_i, \mathbf{K}_i, \mathbf{V}_i &= \mathbf{h}^l\mathbf{W}_i^Q, \mathbf{H}^l\mathbf{W}_i^K, \mathbf{H}^l\mathbf{W}_i^V,\\
	\mathbf{x}_i &= \mathrm{Softmax}(\frac{\mathbf{q}_i\mathbf{K}_i^\top}{\sqrt{d_k}})\mathbf{V}_i,\\
	\mathbf{x} &= \mathrm{Concat}(\mathbf{x}_1, \cdots, \mathbf{x}_h)\mathbf{W}^O,\\
	\mathbf{h}^{l+1} &= \mathrm{ReLU}(\mathbf{x}\mathbf{W}_1)\mathbf{W}_2,\\
	\mathbf{H}^{l+1} &= \mathrm{Copy}(\mathbf{H}^l, \mathbf{h}^{l+1}),
\end{align}
where we lowercase the representations with reduced shape, i.e., $\mathbf{q}_i, \mathbf{x}_i\in\mathbb{R}^{m\times d_k}$, $\mathbf{x}, \mathbf{h}\in\mathbb{R}^{m\times d}$. $d_k$ is the dimension of each attention head. $\mathrm{Copy}(\mathbf{H}^l, \mathbf{h}^{l+1})$ is to copy the hidden states of the exited tokens from $\mathbf{H}^l$ and concatenate with the updated hidden states $\mathbf{h}^{l+1}$. By this token-level early exiting, the computation in self-attention and the following feed-forward network is reduced.

In particular, we show in Table~\ref{tab:flops} the saved MACs (Multiply–Accumulate Operations) in each module of one Transformer encoder layer. We estimate FLOPs with twice the MACs.

\begin{table}[h]
\centering
\resizebox{\linewidth}{!}{
\begin{tabular}{cl|c}
\toprule
\multicolumn{2}{l|}{Module}                                    & Saved MACs \\ \midrule
\multicolumn{1}{l|}{\multirow{4}{*}{SelfAttn}} & LinearProj    & $(n-m)d^2$            \\
\multicolumn{1}{l|}{}                          & MultiHeadAttn & $2n(n-m)(h+d)$             \\
\multicolumn{1}{l|}{}                          & OutProj       & $(n-m)d^2$             \\
\multicolumn{1}{l|}{}                          & LayerNorm     & $2(n-m)d$             \\ \midrule
\multicolumn{1}{l|}{\multirow{2}{*}{FFN}}      & FFN           & $2(n-m)dd_{ff}$             \\
\multicolumn{1}{l|}{}                          & LayerNorm     & $2(n-m)d$             \\ \bottomrule
\end{tabular}
}
\caption{Saved MACs in one Transformer encoder layer. Here we assume $hd_k=d$. $d_{ff}$ is the hidden size of the Feed-Forward Network (FFN) sublayer.}
\label{tab:flops}
\end{table}

\end{document}